\documentclass[11pt,a4paper]{article}
\usepackage[hyperref]{naaclhlt2018}
\usepackage{times}
\usepackage{latexsym}
\usepackage[utf8]{inputenc}
\usepackage{url}
\usepackage{enumitem}
\usepackage{booktabs}
\usepackage{siunitx}

\aclfinalcopy % Uncomment this line for the final submission
\title{Tübingen-Oslo system: Linear regression works the best at Predicting Current and Future Psychological Health from Childhood Essays in the CLPsych 2018 Shared Task}

\author{Çağrı Çöltekin\\
  Department of Linguistics\\
  University of Tübingen, Germany \\
  {\tt \href{mailto:ccoltekin@sfs.uni-tuebingen.de}{ccoltekin@sfs.uni-tuebingen.de}} \\\And
  Taraka Rama \\
  Department of Informatics \\
  University of Oslo, Norway \\
  {\tt \href{mailto:tarakark@ifi.uio.no}{tarakark@ifi.uio.no}} \\
  }

\date{}

\begin{document}
\maketitle
\begin{abstract}
This paper describes our efforts in predicting current and future psychological health from childhood essays within the scope of the CLPsych-2018 Shared Task. We experimented with a number of different models, including recurrent and convolutional networks, Poisson regression, support vector regression, and L\textsubscript{1} and L\textsubscript{2} regularized linear regression. We obtained the best results on the training/development data with L\textsubscript{2} regularized linear regression (ridge regression) which also got the best scores on main metrics in the official testing for task A (predicting psychological health from essays written at the age of 11 years) and task B (predicting later psychological health from essays written at the age of 11).
\end{abstract}

\section{Introduction}

The words we use reflect who we are and how we are.
There have been many successful demonstrations
of predicting personal properties and mental state
from language use of individuals from many forms of linguistic output including social media text.
Examples include 
predicting basic personal features like gender or age 
\cite{barbieri2008,peersman2011,burger2011,nguyen2014}, predicting personality traits
\cite{luyckx2008,celli2013,plank2015}, predicting sentiment towards the topic of a text
\cite{pang2008}, and predicting mental state or health \cite{ramirez2008,coppersmith2014}.
Most of these studies are based on social media text,
and often obtain the target variables (e.g., mental health)
through best-effort approaches based on reports by the subjects,
or the venues where the texts appear.

The Task A of CLPsych-2018 Shared Task \emph{Predicting Current and Future Psychological Health from Childhood Essays} has a similar aim. The task is to predict the mental health of 11 year old children from the essays they have written. The variables to be predicted are depression and anxiety levels as well as a `total' value indicating mental health of the child. The Task B, takes a similar but a more ambitious aim, predicting future mental health from the same essays. More precisely, task B comprises of predicting mental health at ages 23, 33, 42, and 50 given the essays from the age of 11. The task B also includes a surprise component where the mental health at 50 years age is not given in the training set. In Task B, outcome variable is the psychological distress score, based on responses to a questionnaire.
Both tasks are based on a well-known longitudinal study \cite{power2005},
where the psychological health (among other variables) were assessed approximately every ten years after the essays were written.

The problems tackled by both tasks,
assessing current and/or future mental health from linguistic output,
are clearly relevant to monitoring public health,
as well as having possible applications to monitoring or diagnosing
mental health of individuals.
The methods presented can be used
to complement well-established traditional methods,
as well as providing an alternative
where traditional methods are not possible or difficult to employ.
This all may be possible, of course,
if one can achieve reasonable accuracies in these task.

As part of our submission to the CLPsych-2018 Shared Task,
we experimented with a number of different models, both traditional and relatively new,
for predicting the outcome of both tasks. However, we did not attempt to predict age-50 mental health in Task B. Although, we only submitted the results from ridge regression models, we also describe a few other promising models, such as  Poisson regression and recurrent neural networks (RNNs) that we experimented with.

% Write about subtaskA. The task is short term in goal and is about predicting anxiety, depression score, and a total score that is derived from a formula given in pdf. Look up the pdf.
% 
% Write about subtaskB. The second task aims to predict long term health at 23, 33, 42, and 50 years where the target scores for 50 years is not given in the training data.
% 
% We model the task as regression task where we test both traditional regression models such as Ridge regression, Poisson regression (a special case of GLM) with RMSProp, Multi-Output
% 

\section{\label{sec:models}Models}

In this section, we describe the models we experimented with.
Essentially,
we experimented with a number of `traditional' regression models,
and also a few neural network architectures,
which also differ in the way the features are presented to the systems.
Although we were able to experiment with the models discussed here extensively we discuss each model briefly and do not report results with all of these models at this paper.

\subsection{Regression (non-neural) models}

The non-neural models are trained with bag of n-grams as features which are obtained through concatenating the word and character n-grams of different order and weighted through a global sub-linear TF-IDF scaling applied to all the word and character n-grams. We experimented with combinations of different orders of both word and character n-grams.
Besides n-gram order,
we also experimented with case normalization,
and feature selection based on document frequency.

We also explored combination of the control variables (gender and social class) with the textual bag-of-n-gram features.
For the experiments where control variables were included,
gender was coded as a binary input variable,
while we used one-hot (or one-of-k) representation for the social class.

Our submitted system is based on ridge regression,
where the weights were trained to minimize the L\textsubscript{2} regularized sum of squared error.
We log transformed the outcome variable,
which was shifted linearly with a constant (to avoid $\log 0$) before transformation.
We applied the inverse transformation at prediction time.
In the case of both task A and task B,
we treat all the three target variables as outcomes in a single linear regression model.
The regularization parameter $\alpha_j$ was tuned separately for each target variable ($1<=j<=3$)
through cross-validation by searching for the best combination of n-gram orders and $\alpha$ parameter.

In addition to linear regression, we also experimented with Poisson regression for task A. The intuition behind the experiments with Poisson Regression is that the target variable in task A is similar to count data. Therefore, we assume that a target variable such as `total'  variable is drawn from an independent Poisson random variable with mean $\lambda$. For a document feature vector $x_i$, the mean $\lambda_i$ is equal to $\exp(\theta^T \mathbf{x_i})$. Subsequently, the probability of observing the target variable under a Poisson variable with mean $\lambda_i$ is given by Poisson distribution. The Poisson regression model is a special case of Generalized Linear Model \citep{neldergeneralized1989}.
The model is trained through applying Stochastic Gradient Descent with RMSProp algorithm \citep{tran2015stochastic}. However, we found that the results of the Poisson regression model were not better than the Ridge regression model at task A.

Besides being `count data',
another property of the outcome variables in this task is
that the outcome variables is zero for many data points.
This type of data is modeled better through so-called zero-inflated models \cite{lambert1992},
which in essence a two-stage model where the data points are classified as zero and non-zero,
and a regression model, e.g., Poisson regression, is applied only to data points predicted to be non-zero. 
We also tried a two-stage model along these lines.
However, the initial results were rather discouraging and we did not experiment with this model thoroughly.

We also explored a support vector regression model for task A and task B which is trained in a similar fashion to SVM with TF-IDF features. The model's performance was close but lower than the L\textsubscript{2} regularized linear regression and therefore, we do not report the results of the model.

We also experimented with random forest regression and Bayesian regression both of which have been shown to be useful in a number of similar tasks. However, the implementations we have access to were not scalable due to the dimensionality of TF-IDF vectors and the computation time required. Hence, we do not report any results with these two methods either. The models discussed in this section were implemented with Scikit-learn package \citep{sklearn} using liblinear back end \citep{liblinear}. The Poisson regression model was implemented using Numpy.

\subsection{Neural networks}
In this paper, we experimented with a neural model consisting of bidirectional Gated Recurrent Units with character and word embeddings trained for the task. The first layer of the model consists of a separate embedding layers built on characters and words. The concatenated output vectors from character and word embeddings are then supplied as input to a Gated Recurrent Network \citep{cho2014}. The length of sequence was fixed at 1500 characters for training character embeddings and at 400 words for training word embeddings. The documents are lowercased for training words and characters embeddings. The number of GRU units was also fixed to reflect the sequence length in the case of GRU units. The output of the GRU network had a dimension of 256 and is followed by a fully connected layer with a single output that outputs a real number. The network is trained using Adam optimizer with mean squared error as the objective. All the neural models are implemented using Keras \citep{chollet2015keras} with Tensorflow as the backend \citep{tensorflow}.

\section{Experiments and results}
In this section, we describe the dataset, methods, experiments, and results.

\subsection{Data}

The data for the shared task comes from National Child Development Study (NCDS),
which is a longitudinal study following 17,416 babies born in Britain in 1958,
\cite{power2005}. The part of the study relevant to the present task is
the essays written by a subset of children at age 11. The shared task training data includes 9217 essays.
We used only the training data for which none of the variables were missing. After removing the training instances with missing target variables, we were left with 9146 instances for Task A and 4938 instances for Task B. The training documents used for Task B is
a subset of the training documents used for Task A.
The length of the documents used for Task A have mean of 964.56 characters and 227.19 words.
The document length exhibits quite some variability with standard deviations of 503.07 and 116.52 respectively.

Besides essays, the data includes two background control variables: gender, and the social class at age 11. We also used these variables as inputs to our models. The outcome variables for Task A are scores
(number of underlined sentences) indicating anxiety,
depression and total score (number of sentences underlined).
The outcome variables for Task B are the number of 
questions indicating psychological distress.
An important observation for all outcome variables is
that the distributions are heavily skewed with many zero values.

\subsection{Evaluation metrics}
The predictions of the models are evaluated using mean absolute error and disattentuated R.%
\footnote{\url{https://www.rasch.org/rmt/rmt101g.htm}}
The systems are ranked using disattentuated R which is a modification to Pearson's R,
to account for the correlation between the outcome variables.

\subsection{Experimental procedure and hyperparameter tuning}

We did not do extensive parameter tuning with all the models section \ref{sec:models}.
After some initial experiments,
ridge regression and support vector regression seemed to yield promising performance scores.%
\footnote{To put it another way,
  the others, e.g., neural networks,
  we initially expected to perform better did not yield expected results.}
Hence, we run a random search through the following hyperparameter values.

\begin{description}
      \setlength{\itemsep}{0.2ex}
  \item[\normalfont\texttt{c\_ngmax}] Maximum character n-gram order: 1--8
  \item[\normalfont\texttt{w\_ngmax}] Maximum word n-gram order: 1--5
  \item[\normalfont\texttt{min\_df}] Document frequency cutoff for feature selection: 1--5
  \item[\normalfont\texttt{$\alpha$}] Regularization constant ($\alpha$): 0.5--20.0 
    (we used $1/\alpha$ for SVR margin parameter C)
  \item[\normalfont\texttt{lowercase}] Case normalization: character n-grams, word n-grams, both or none
  \item[\normalfont\texttt{ctrl\_weight}] Weight of control variables: 0.0--1.0
  \item[\normalfont\texttt{a11\_weight}] Weight of Age-11 predictions: 0.0--1.0 (Task B only)
\end{description}

We used 5-fold cross validation on the training set for determining the best hyperparameter configuration. We trained the model with the best hyperparameters on the complete training set
before producing the final predictions.

\subsubsection{Task A}

For Task A, we obtained best results on training set using 5-fold cross validation
using the hyperparameter configuration reported in table \ref{tbl:paramA}.
We obtained best disattentuated R scores of  0.5778, 0.2315 and 0.4678
for total, anxiety and depression respectively
on training set with the parameter values in table~\ref{tbl:paramA}.
Similar performance scores were obtained using other (rather diverse) parameter settings.
From a quick inspection,
we did not observe any clear trends regarding usefulness of parameter values,
except more features (higher \texttt{c\_ngmax} and \texttt{w\_ngmax} values seem to help).

This model yielded disattentuated R scores of 0.5788, 0.153 and 0.4669
for total, anxiety and depression on the training set respectively.
This model also obtained the top rank (based on total score)
in Task A among other shared task participants.

\begin{table}
  \begin{center}
    \begin{tabular}{lSS}
      \toprule
      Hyperparameter & {Ridge} & {SVR}\\
      \midrule
      \texttt{c\_ngmax} & 5 & 6\\
      \texttt{w\_ngmax} & 3 & 2\\
      \texttt{min\_df} & 2 & 1\\
      \texttt{lowercase} & {word} & {word}\\
      $\alpha_\texttt{total}$ & 5.0 & 5.0\\
      $\alpha_\texttt{anxiety}$ & 5.0 & 10.0\\
      $\alpha_\texttt{depression}$ & 5.0& 20.0\\
      \texttt{ctrl\_weight} & 0.5 & 0.5 \\
      \bottomrule
    \end{tabular}
  \end{center}
  \caption{\label{tbl:paramA}%
    Best hyperparameter values for ridge regression (Ridge)
    and support vector regression (SVR) models for Task A.
    The values are obtained through a random search 
    from approximately 400 random parameter settings.
  }
\end{table}

% A: 'c_ngmax=5;w_ngmax=3;alpha=(5.0,5.0,5.0);mix_ratio=0.5;shift=2;min_df=2'                                                                                                       B: 'c_ngmax=4;w_ngmax=5;alpha=(3.0,8.0,10.0);mix_a11=0.0;mix_ctrl=1.0'                                                                                                            

%\textcolor{red}{Add a figure of n-gram orders}

\subsubsection{Task B}

The best parameter settings for both ridge regression and support vector regression models
for Task B are reported in table~\ref{tbl:paramB}.
The ridge regression model with the hyperparameter settings reported in table~\ref{tbl:paramB}
obtained psychological distress correlations (disattentuated) of 0.4118, 0.2919 and 0.2527
for ages 23, 33 and 42, respectively.
The inclusion of age-11 predictions as predictors in Task B was useful
if gold-standard scores were used.
However, we did not observe any benefits if predicted age-11 outcomes were used.
As a result,
we used the same model,
except we did not use the predicted age-11 scores as predictors in our final model.
The model obtained disattentuated R scores of 0.443, 0.3175 and 0.1961
for ages 23, 33 and 42, respectively, on the official evaluation.
With an average of 0.3189, it ranked best among other participating models.

\begin{table}
  \begin{center}
    \begin{tabular}{lSS}
      \toprule
      Hyperparameter & {Ridge} & {SVR}\\
      \midrule
      \texttt{c\_ngmax}               & 4 & 7\\
      \texttt{w\_ngmax}               & 5 & 5\\
      \texttt{min\_df}                & 1 & 1\\
      \texttt{lowercase}              & {word} & {word}\\
      $\alpha_\texttt{total}$         & 3.0 & 8.0\\
      $\alpha_\texttt{anxiety}$       & 8.0 & 20.0\\
      $\alpha_\texttt{depression}$    & 10.0& 20.0\\
      \texttt{ctrl\_weight}           & 1.0 & 0.1 \\
      \texttt{a11\_weight}            & 0.5 & 0.1 \\
      \bottomrule
    \end{tabular}
  \end{center}
  \caption{\label{tbl:paramB}%
    Best hyperparameter values for ridge regression (Ridge)
    and support vector regression (SVR) models for Task B.
    The values are obtained through a random search 
    from approximately 400 random parameter settings.
    \texttt{a11\_weight} is based on predicted age-11 outcomes.
  }
\end{table}

Similar to Task A,
the support vector regression model yielded similar but again slightly lower results.
We obtained disattentuated R scores of 0.4092, 0.2779 and 0.215 on the training set
with the parameter values presented in table~\ref{tbl:paramB}.

%\textcolor{red}{Add a figure of n-gram orders}

\section{Discussion}

In this paper we described the models we experimented with in our participation of
CLPsych-2018 Shared Task, and reported our results.
For our submission, 
we used a ridge regression model with bag-of-n-gram features as our final model.
The model we used is simple linear regression with L$_2$ regularization,
except the outcome variable was log-transformed.
The model obtained best results on both tasks at official evaluation.
We have also experimented with a range of other models,
including Poisson regression and neural models.
However, based on (somewhat) limited tuning efforts,
none of these systems achieved scores close to ridge regression and support vector regression models
(where the SVR model was close to the results of ridge regression model).

The promise of this line of work, namely, predicting mental health from language samples is interesting scientifically and it may also have important applications in monitoring public and personal health.
Predicting future mental health is even more interesting as it may allow the clinicians to identify preventive interventions.
The method is even more relevant and easily applicable due to the increase in the longitudinal collection of language output in the last few decades.,
%collecting samples language output has become increasingly easier during the last few decades.
These prospects are only attainable if we can predict present or future mental health
from language samples with reasonable accuracy.
Our results show that there is in fact a rather strong signal in the language samples for detecting mental health.
As expected,
the predictions are less reliable as time gap between the language output
and the prediction is wider.
However, 
there is small but reliable correlation between language output at age 11 and psychological health at age 42.

The correlation results of both this paper and other participants clearly show that
the linguistic data contains cues that can predict current and future mental health.
One of the interesting questions is whether we can get more information from this data. 
To this end, better modeling of the data may allow us to get more information,
i.e., detect more of the signal within the text at hand. With models involving Poisson regression, zero-inflated models, and neural networks, we tried to use models that are more suitable for the data at hand. However, we obtained best results with relatively simple linear models applied to log transformed data. The success of the simple linear models over more complex (e.g., neural networks) is
in line with our experiences in a diverse set of text classification tasks
\cite{coltekin2016,rama2017,coltekin2018}.
The results indicating superiority of simple linear models, however,
should not be considered as conclusive since exploration of more complex models was not thoroughly performed due to time limitations.

% \section*{Acknowledgments}
% 
% The acknowledgments should go immediately before the references.  Do
% not number the acknowledgments section. Do not include this section
% when submitting your paper for review. \\

\bibliographystyle{acl_natbib}
\bibliography{clpsych2018}

\end{document}